\documentclass[10pt,twocolumn,letterpaper]{article}

\usepackage{iccv}
\usepackage{times}
\usepackage{epsfig}
\usepackage{graphicx}
\graphicspath{{.}}
\usepackage{amsmath}
\usepackage{amssymb}
\usepackage{array}
\usepackage{tabulary}
\usepackage{dblfloatfix}
\usepackage{booktabs}
\usepackage{tikz}
\usepackage{todonotes}
\usepackage{flushend}
\usepackage{caption}
\usetikzlibrary{arrows}
\usetikzlibrary{quotes}
\usetikzlibrary{patterns}
\usetikzlibrary{shapes.geometric}
\tikzstyle{arrow} = [thick,->,>=stealth]
\tikzstyle{encoding} = [rectangle, rounded corners, minimum width=1.cm, minimum height=0.3cm, text centered, draw=black, align=center]
\tikzstyle{input} = [rectangle, minimum width=1.cm, minimum height=0.3cm, text centered, align=center]
\tikzset{font={\fontsize{6pt}{12}\selectfont}}
\tikzset{
  every node/.style={
  , execute at begin node=\setlength{\baselineskip}{1em}
  }
}
\newcommand{\trajfc}[3]{
\node[encoding, #2] (#1) {FC 256 \\ FC 64 \\ FC 10};
\node[below of = #1] (#1_out) {Trajectory};
\node[below of = #1_out, yshift=0.75cm] {#3};
\draw[arrow] (#1) -> (#1_out);
}

\newcommand{\netin}[3]{
\node[input, above of=#3] (#1) {#2};
\draw[arrow] (#1.south) -> (#3.north);
}

\newcommand{\concatlayer}[2]{
\node[circle, #2, draw=black] (#1) {$\cup$};
}

\usepackage[pagebackref=true,breaklinks=true,letterpaper=true,colorlinks,bookmarks=false]{hyperref}

\iccvfinalcopy 

\def\iccvPaperID{1} 

\ificcvfinal\fi
\begin{document}

\title{Conditional Vehicle Trajectories Prediction in CARLA Urban Environment}
\author{Thibault Buhet, Emilie Wirbel,  Xavier Perrotton\\
Valeo Driving Assistance Research\\
34 rue Saint Andr\'e, 93000 Bobigny\\
{\tt\small name.surname@valeo.com}
}

\maketitle

\begin{abstract}
Imitation learning is becoming more and more successful for autonomous driving. End-to-end (raw signal to command) performs well on relatively simple tasks (lane keeping and navigation). Mid-to-mid (environment abstraction to mid-level trajectory representation) or direct perception (raw signal to performance) approaches strive to handle more complex, real life environment and tasks (e.g. complex intersection). In this work, we show that complex urban situations can be handled with raw signal input and mid-level representation. We build a hybrid end-to-mid approach predicting trajectories for neighbor vehicles and for the ego vehicle with a conditional navigation goal. We propose an original architecture inspired from social pooling LSTM taking low and mid level data as input and producing trajectories as polynomials of time. We introduce a label augmentation mechanism to get the level of generalization that is required to control a vehicle. The performance is evaluated on CARLA 0.8 benchmark, showing significant improvements over previously published state of the art.
\end{abstract}

\section{Introduction}

Modular pipelines \cite{thrun2006stanley} are the most used approach to autonomous driving. The advantage is that the modules are interpretable and relatively mature, in particular on the perception side with the success of deep learning for object detection (\cite{he2017mask, liu2016ssd} among many others). However, the complexity of the interactions in the real world causes the pipeline to be also complex, especially in the planning and decision modules. The annotations for the modules are also costly and difficult to obtain.

End-to-end imitation learning (IL) is a possible answer to these issues: one single neural network is used from the raw data input to the command output, and the ground truth command is obtained by recording an expert without additional annotation effort. Note that the expert does not necessarily have to be a human: Pan \etal \cite{pan2018agile} use a Model Predictive Control agent with high end sensors to train for aggressive offroad driving. One of the main difficulties of IL is that the online test distribution is not the same as the recorded ground truth. This is due in particular to error accumulation from the network, which will lead the vehicle away from an ideal trajectory, where there is almost no data recorded because an expert driver avoids these situations. Without data augmentation, the common offline evaluation metrics are not correlated to the online performance, as shown for example by Codevilla \etal \cite{codevilla2018offline} who stress the importance of data augmentation and study offline metrics. The two seminal articles for IL in autonomous driving \cite{pomerleau1989alvinn, bojarski2016end} propose a solution: combine side cameras with the main one to emulate lateral deviations from the road. In \cite{toromanoff2018end, george2018imitation} the recorded data is modified to perform \textit{label augmentation}: data similar to failure cases is generated a posteriori, but without the need of additional sensors. This augmentation is usually necessary to use IL directly, unless some additional data is collected online iteratively to correct the failure cases, using DAgger \cite{ross2011reduction} for example.

Lack of interpretability is another challenge of IL, or more generally making sure the network uses the correct information. \textit{Direct perception} is a possible solution: instead of commands, the network predicts hand-picked parameters relevant to the driving (distance to the lines, to other vehicles), which are then fed to an independent controller \cite{chen2015deepdriving, al2017deep, sauer2018conditional}. The limitation of this approach is the necessity to choose the relevant parameters.

Another option is to use mid level interpretable data as input and output of the network: the input is coming from perception modules, the output is a mid level representation like a trajectory. ChauffeurNet \cite{bansal2018chauffeurnet} is an example of this \textit{mid-to-mid} approach: road users positions, road geometry and traffic light states are used to produce vehicle trajectories and control a real vehicle. This work also demonstrates that adding high level information makes it possible to scale up in terms of complexity of the situations.

\textit{Priviledged learning} is another possible way to improve the network performance by providing additional information. In that approach, the network is partly trained with an auxiliary task on a ground truth which is useful to driving, and on the rest is only trained for IL. The goal is to leverage the fact that part of the network layers were trained to produce features which are related to helpful information. Xu \etal \cite{xu2017end} use semantic segmentation on the Berkeley Deep Drive dataset as an auxiliary task, and show that it increases performance on the future ego-motion prediction tasks.

\begin{figure}[h]
\centering
\includegraphics[width=0.8\columnwidth]{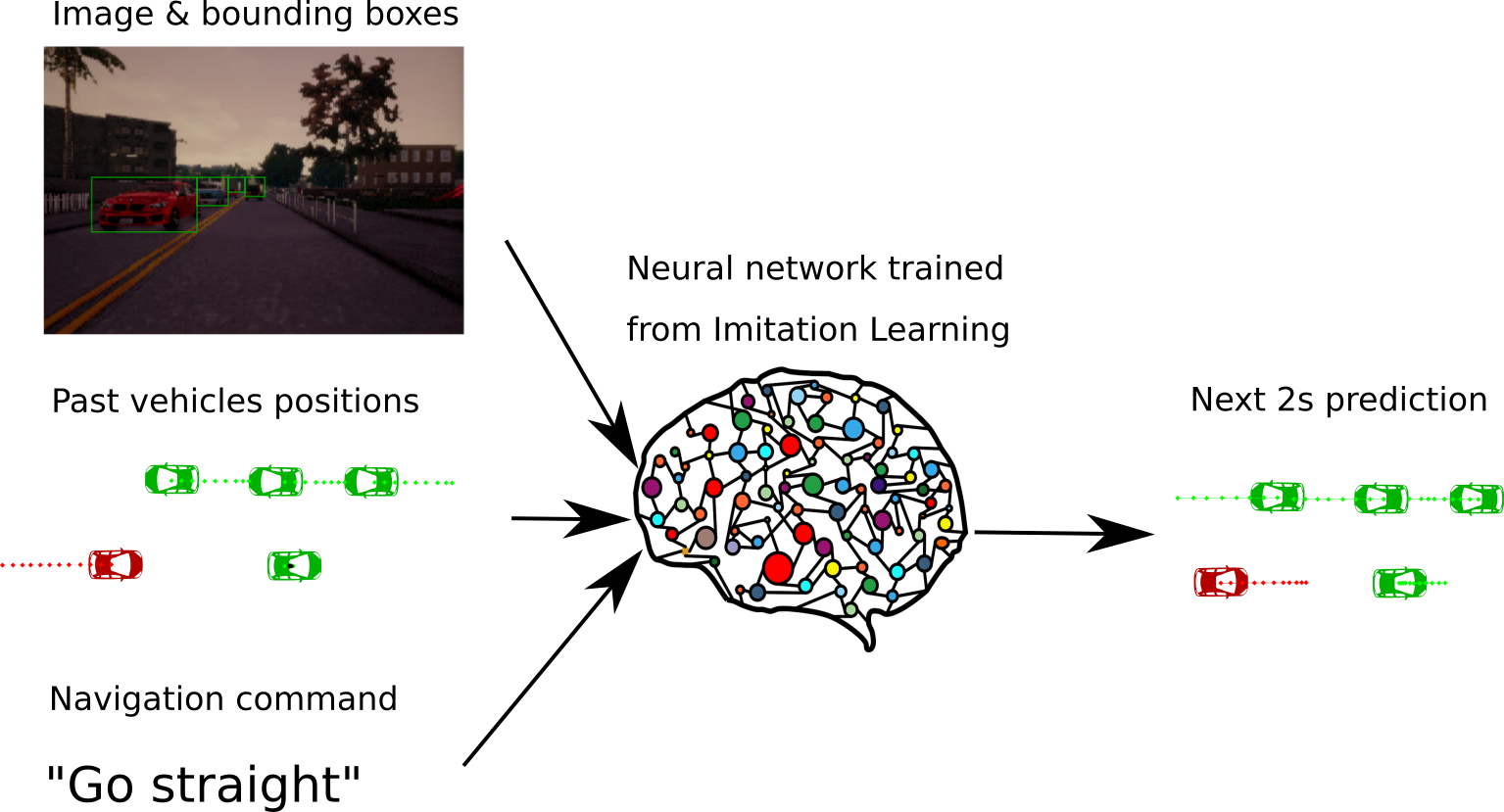}
 \caption{Inputs and outputs of the network: image, object tracks and navigation command, to predict ego and neighbors trajectories }
 \label{fig:gen_structure}
\end{figure}

In this article, our goal is to perform navigation tasks in CARLA \cite{dosovitskiy2017carla} of an autonomous vehicle (\textit{ego vehicle}), while also predicting the trajectories of external agents in the surroundings (\textit{neighbors}). To do so, we propose a network taking as input: images, object detections as 3D and 2D bounding boxes and navigation commands to describe the desired behavior of the ego vehicle at intersections. It is trained to predict ego vehicle and neighbors trajectories through imitation learning, expressed as polynomials over time. This is a hybrid \textit{end-to-mid} approach, because both raw signal and partial environment abstraction are used, and trajectories are produced. It combines the approaches of end-to-end, mid-to-mid and priviledged learning Our main contributions are:
\begin{itemize}
 \item to propose an architecture combining raw image data and mid-level information. Our network uses contributions from pure trajectory prediction (see section~\ref{sec:related_work} for more details), but also raw camera image to predict the desired output, typically for traffic lights.
 \item to introduce an augmentation method for these vehicle trajectories
 \item to use neighboring vehicles trajectory prediction as a simultaneous task.
 \item to perform an extensive evaluation on the CARLA 0.8 benchmark for the control of a vehicle using our trajectory prediction, with a comparison to state of the art baseline and ablation study of the augmentation and neighbors prediction
\end{itemize}

\section{Related work}
\label{sec:related_work}

\paragraph{Conditional Imitation learning for autonomous driving} Since our goal is to perform navigation but not path planning, we use navigation commands as an input. To take them into account, they can be simply concatenated in the last layers of the network \cite{hubschneider2017adding}, but this requires parameterization to give them an appropriate weight. In Conditional Imitation Learning (CIL) \cite{codevilla2018end}, the last layers of the network are split into branches which are masked with the current navigation command, thus allowing the network to learn specific behaviors for each goal. This method is used as is in this work because it is more scalable and interpretable. Conditional Affordance Learning (CAL) \cite{sauer2018conditional} adapts this to direct perception, and improves over CIL. CIL and CAL will be used as baselines for evaluation.

\paragraph{Simulator and benchmark} A simulator is an ideal environment for IL: it is easy to generate training data, to add additional sensors for data augmentation, and to get quantitative evaluation on the online test phase. CARLA
\cite{dosovitskiy2017carla} is used for this article: it is dedicated to autonomous driving, with a benchmark which is used to compare to the state of the art (see section~\ref{sec:results}). Other simulators are available for autonomous driving research applications. DeepDrive \cite{craig_quiter_2018_1248998} is a simulator originally based on the game Grand Theft Auto V (GTA V). GTA V itself has been used as a realistic simulator, but lacks flexibility for imitation learning. At the time this article is completed, a new simulator, LGSVL Simulator \cite{lgsvl2019simu}, has just been released with promising features (large environment and sensor set) but still lacks some features like autopilot for imitation learning.

\paragraph{Predicting trajectories and interactions from object positions} Many existing contributions cover future trajectory prediction from past positions, both model based trajectory prediction \cite{mouhagir2017using} and with Recurrent Neural Networks (RNN) \cite{park2018sequence}. SocialLSTM \cite{alahi2016social} leverages RNNs to model interactions, here between pedestrians. SocialGAN \cite{gupta2018social} uses GAN to generate pedestrians trajectories with plausible interactions. The common idea is that interactions between agents are learnt by sharing encodings of the individual trajectories which are near to each other. The individual encodings of the trajectories are generally produced by shared layers: all agents are interchangeable. \cite{deo2018convolutional} shows an extension of the concept to autonomous vehicles, using a proximity map (similar to an occupancy grid) and separate trajectory encoding. Our article extends the concept of trajectory prediction with polynomials, and proves that it can be combined with image level encoding and navigation commands.

\paragraph{Predicting trajectories from low level data} Predicting object trajectories from raw data has already received some attention. Huang \etal \cite{huang2016deep} perform visual path prediction from a fixed surveillance camera, to predict potential trajectories from pedestrians and cars, building a reward map of reachability. In contrast, our work relies on a moving camera depending on the ego vehicle. \cite{lee2017desire, rhinehart2018deep} use Lidar information in combination with image, either with imitation or model based reinforcement learning. Luo \etal \cite{luo2018fast} use an end-to-end network based on 3D Lidar data to track and predict future motion. We show in our work that motion forecasting can be performed with only 3D detections and 2D image information. Intention-Net \cite{wei2017intention} combines raw image with a high level plan where the desired path is drawn for indoor navigation, which shows that it is possible to take advantage of metric structure information and low level image data. This is similar to what is done in this article, but here the network itself predicts the future path.

\section{Trajectory prediction from image and object detections}

\subsection{System description}

The goal of the system is to predict the future positions on a fixed 2s horizon of the ego vehicle and the neighbors. The maximum number of predicted neighbors is fixed at $N=5$ for simplicity. It is assumed that the past positions of the neighbors are known relatively to the current ego position (here obtained from the simulator). For the ego vehicle, the future trajectory is conditionned by a navigation command. This future trajectory can be fed to a controller, here a Proportional Integral Derivative (PID). All experiments described in this paper are conducted in CARLA, but could be extended to any system providing object tracks in 3D with a 2D reprojection, raw image input and ego vehicle odometry. CARLA autopilot is used as the expert.

To describe the trajectory, we represent positions of the object as tuples $(x(t), y(t))$ for time $t$, corresponding to the center of the 3D bounding boxes for the neighbors. To describe these efficiently, we chose to use a polynomial representation. For each predicted trajectory, we predict the vector $(x_0, ..., x_4, y_0, ..., y_4)$ where

\begin{align*}
 x(t) &= x_0 t^4 + x_1 t^3 + x_2 t^2 + x_3 t + x_4 \\
 y(t) &= y_0 t^4 + y_1 t^3 + y_2 t^2 + y_3 t + y_4
\end{align*}

Note that the coefficients are proportional to speed, acceleration, jerk and derivative of jerk along the x and y axis. For the other vehicles, the trajectory is given relatively to their current position. This makes it possible to predict vehicle trajectories independently of their distance to the ego vehicle. This is a valid assumption because these shifted trajectories have the same shape as the original ones, and the influence of the relative vehicle positions is still encoded in the proximity map (see Section~\ref{sec:network_structure} for details). This is also a way to get more relevant training data and generalization, because all inputs are in a similar domain.

\subsection{Network structure}
\label{sec:network_structure}

The following data is fed as input to the network:
\begin{itemize}
    \item a front facing camera RGB image $I$ ($320 \times 240 \times 3$)
    \item the past positions of the ego vehicle and neighbors, sampled every $\delta t = 0.1s$ for the last 2s, which corresponds to $T=20$ values. They are given in the reference of the current ego vehicle position. They are grouped using a sliding window of size $K=3$.
    These vectors are noted $V_i$ for neighbors (with $i \in [0, N[$) and $E$ for the ego vehicle.
    \item a navigation command $nc$. If we are approaching an intersection, the goal can be \textit{left}, \textit{right} or \textit{cross}, else the goal is \textit{keep lane}.
\end{itemize}

$V_i$ and $E$ are projected into a proximity map $M$. This map represents a local neighborhood of the current ego vehicle position, of size $65 \times 10.5$m, with the current ego at the center. The dimensions of the area around the ego, used in \cite{deo2018convolutional}, corresponds to 3 standard lanes side by side for around 2s front and rear at 50kph. We use a coarse resolution ($13 \times 3$) to represent this neighborhood as an array where each cell is roughly the size of a vehicle. Each cell contains K=3 consecutive positions ending at the corresponding instant, if there is a vehicle at that position and at that time, and nothing elsewise. We stack this representation for the past $T$ frames, resulting in a $13 \times 3 \times T \times 2 \times K$ array. Note that the information of $V_i$ is contained in $M$: we distinguish the two to be able to take an arbitrary neighbors into account for the context, but still be able to predict the future of $K$ chosen tracks.
The past 3D bounding boxes of the road users in the current reference are projected back in the current camera space. This gives a $320 \times 240 \times K$ array $B$, where the last dimensions contains the bounding box at t=0 plus the 2 previous ones.

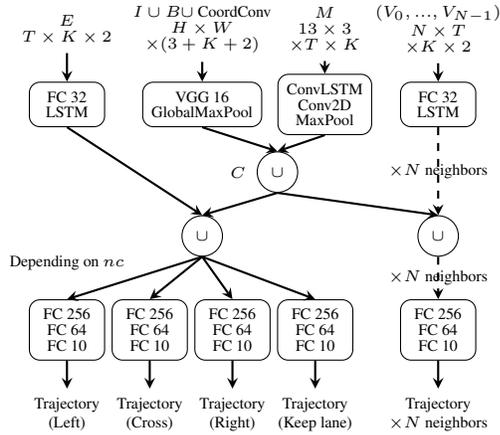
\begin{figure}[h]
    \centering
    \begin{tikzpicture}[node distance=1cm]
     \node[encoding] (egoenc) {FC 32 \\ LSTM};
     \node[encoding, left of=egoenc, anchor=west, xshift=2cm] (imenc) { VGG 16 \\ GlobalMaxPool};
     \node[encoding, left of=imenc, anchor=west, xshift=2cm] (proxyenc) {ConvLSTM \\ Conv2D \\ MaxPool};
     \node[encoding, left of=proxyenc, anchor=west, xshift=2cm] (neighbenc) {FC 32 \\ LSTM};
     \netin{egoin}{$E$ \\  $T \times K \times 2$}{egoenc}
     \netin{im_in}{$I \cup B \cup$ CoordConv \\ $H \times W$ \\ $ \times (3 + K + 2)$}{imenc}
     \netin{map_in}{$M$ \\ $13 \times 3$ \\ $\times  T \times K$}{proxyenc}
     \netin{neig_in}{$(V_0, ..., V_{N-1})$ \\ $N \times T$ \\ $ \times K \times 2$}{neighbenc}
     \concatlayer{context}{below of=imenc, right of=imenc, yshift=0.1cm}
     \node[left of=context, anchor=west, xshift=0.25cm] {$C$};
     \concatlayer{egoconcat}{below of = imenc, yshift=-0.75cm}
     \concatlayer{neighbconcat}{below of = neighbenc, yshift=-0.75cm}
     \draw[arrow] (context.south) -> (egoconcat.north);
     \draw[arrow] (context.south) -> (neighbconcat.north);
     \draw[arrow] (imenc.south) -> (context.north);
     \draw[arrow] (proxyenc.south) -> (context.north);
     \draw[arrow] (egoenc.south) -> (egoconcat.north);
     \draw[arrow, dashed] (neighbenc.south) -- node {$\times N$ neighbors} (neighbconcat.north);
     \trajfc{lefttraj}{below of = egoenc, yshift=-2cm}{(Left)}
     \trajfc{crosstraj}{right of = lefttraj, xshift=0.1cm}{(Cross)}
     \trajfc{righttraj}{right of = crosstraj, xshift=0.1cm}{(Right)}
     \trajfc{keeptraj}{right of = righttraj, xshift=0.1cm}{(Keep lane)}
     \draw[arrow] (egoconcat.south) -> (lefttraj.north);
     \draw[arrow] (egoconcat.south) ->(crosstraj.north);
     \draw[arrow] (egoconcat.south) -> (righttraj.north);
     \draw[arrow] (egoconcat.south) -> (keeptraj.north);
     \node[above of = lefttraj, yshift=-0.1cm] { Depending on $nc$};
     \trajfc{neighbtraj}{below of = neighbconcat, yshift=-0.25cm}{$\times N$ neighbors}
     \draw[arrow, dashed] (neighbconcat) -- node {$\times N$ neighbors} (neighbtraj);
    \end{tikzpicture}

    \caption{Trajectory prediction network (FC \textit{X}= fully connected with \textit{X} outputs, Conv2D=regular convolution, $\cup$=Concatenation, see section~\ref{sec:network_structure} for other notations)}
    \label{fig:net_structure}
\end{figure}

%
%
  

The full structure of the network is described in Figure~\ref{fig:net_structure}. The global context from both image and proximity map is encoded into two feature vectors, concatenated to form a context feature vector $C$. The image concactenated with the neighbors bounding boxes is encoded using a VGG16 \cite{simonyan2014very} network structure in which we replaced the first convolutional layer by a CoordConv \cite{liu2018intriguing} layer with the same parameters, followed by a global max pooling layer to make the feature vector dimension independent from the input dimension. The VGG16 could be replaced by any other image encoder, but makes visual backpropagation \cite{bojarski2018visualbackprop} easier. The proximity map is encoded using a Convolutional Long Short Term Memory layer (ConvLSTM) \cite{xingjian2015convolutional}, followed by a convolution and a max pool. Note that here the proximity maps are treated as images and not as a time serie.

In parallel, the ego vehicle trajectory and the other vehicles trajectories are encoded, with a fully connected layer followed by an LSTM \cite{hochreiter1997long}. The neighbors encoding are $N$ twin layers with shared weights. The trajectory encodings are concatenated with the context encoding, then fed into fully connected layers to finally produce the polynom coefficients. The neighbor weights are shared between neighbors and separate from the ego weights.

Finally, the context encoding is concatenated with the ego encoding, then fed into branches corresponding to the different high level goals. A mask is used to choose the branch corresponding to the current high level goal. Neighbors encodings are concatenated with the context encodings then decoded in parallel, with shared weights.

As stated in the introduction, this network structure is a significant contribution on pre-existing work \cite{codevilla2018end, deo2018convolutional}. Part of it relies on the principles of Social LSTM and the proximity map of \cite{deo2018convolutional}, but with an original contribution on the trajectories decoding, the final output format and the distinction between the controlled ego vehicle and the neighbors. It also introduces the concept of global image context encoding concatenation, which is then decoded individually.

\section{Training the network}

\subsection{Ground truth and loss}

To generate the ground truth, scenarios are recorded with different vehicles driving around the CARLA Town~01, one of them being the ego vehicle. Once the data is recorded, training data is extracted by using a sliding window of size 4s: 2s for the past, and 2s for the anticipation. On each of these windows, the positions of the ego vehicle and its neighbors are computed relatively to the middle of the window. This makes it possible to generate $E$, $M$ and $(V_0, ..., V_{N-1})$. The input image $I$ is the camera frame at the middle of the time window. All positions of the neighbors vehicles are projected back into $I$, using the camera calibration, so that $B$ can be generated. Least square polynomial fit is used to get the coefficients for the future trajectory, based on the points from the second half of the window. 
The navigation command is computed offline using the recorded positions and orientation of the ego vehicle and a list of positions of all the intersections in the map.

The loss is not done directly on the estimated polynomials coefficients, to be independent from the polynom degree and the degree of the coefficient. For example, an error on coefficient $x_0$ has much more impact than on $x_4$. Instead, 2D points of the fixed anticipation are sampled with the time step $\delta t$, which are noted $e_i$ and $\hat{e}_i$ with $i \in [1, T]$ for the ego vehicle ground truth and prediction respectively, and $v_{k, i}$ and $\hat{v}_{k, i}$ for the $k$-th neighbor ground truth and prediction respectively. We then use a L2 loss on these points:

\begin{align*}
 L(\hat{e}_i, e_i, \hat{v_{k, i}}, v_{k, i })= \sum_{i = 1}^{T} || \hat{e}_i - e_i ||^2 + \sum_{k = 0}^{N} \sum_{i = 1}^{T} || \hat{v}_{k, i} - v_{k, i} ||^2
\end{align*}

For training, we used Adam optimizer with a starting learning rate of $10^{-5}$ and a batch size of 8.

\subsection{Label augmentation}

The main challenge of IL is the difference between train and online test distribution, in particular since the ego vehicle is controlled with the trajectory prediction. To reach the acceptable level of performance, we combine classical randomization with label augmentation, which means generating new inputs and labels from existing data.

Randomization is added first to ensure a better generalization. To increase the variability in the training data, we successively recorded 3 minutes episodes with randomized parameters (40-80 vehicles, 10-30 pedestrians, starting positions). In addition, all vehicles models and pedestrians models are chosen randomly, and the paths of the vehicles and the pedestrians are also random. The map used, Town~01, is fixed by the benchmark training conditions and the weather is random between the 4 weathers allowed for training: clear noon, wet noon, hard rain noon, clear sunset.

Noise is added to the input observations, to avoid overfitting, ensure better generalization but also emulate noisy data that could be obtained from a real environment. It also prevents the network from learning to simply extrapolate from the past trajectory, as noted for example in \cite{bansal2018chauffeurnet}. A Gaussian noise is added to the past positions and image bounding boxes of the vehicles. In parallel, vehicles are randomly removed from or added to the proximity map.

\begin{figure}[t]
\centering
 \includegraphics[width=0.75\columnwidth]{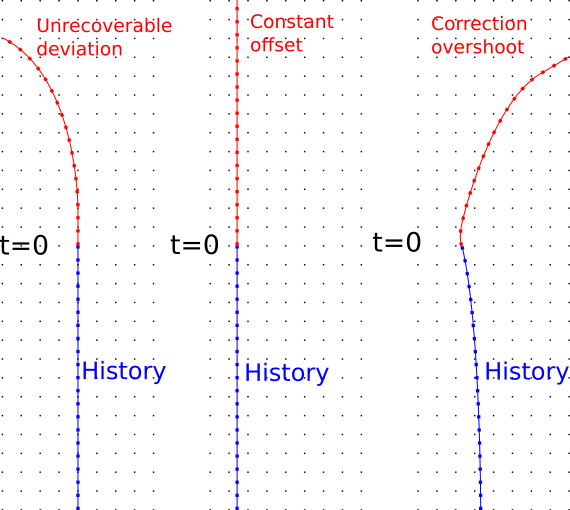}
 \caption{Example of problematic behavior produced without label augmentation: unrecoverable deviation (left), lateral offset (center) and recovery overshoot (right)}
 \label{fig:bad_behav}
\end{figure}

As expected, a training without label augmentation to prepare online behavior yields poor results because of error accumulation. After training and testing online without label augmentation three main problematic behaviors were observed (represented in Figure~\ref{fig:bad_behav}): \textit{unrecoverable deviation}, \textit{lateral offset} (the prediction has a constant lateral offset relatively to the ground truth), and \textit{recovery overshoot} (the prediction over-compensates after a deviation).


\begin{figure}[t]
\centering
 \includegraphics[width=0.75\columnwidth]{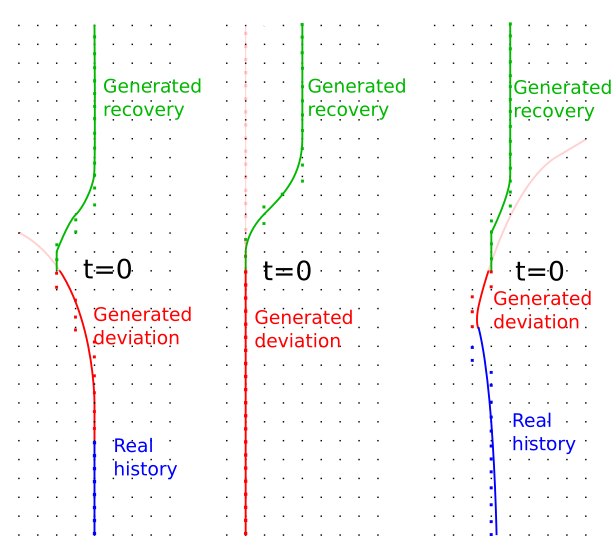}
 \caption{Example of label augmentation for incorrect behavior correction: large error recovery (left), lane recentering (center) and overshoot compensation (right)}
 \label{fig:cor_beh}
\end{figure}

To correct these behaviors, we create a randomized artificial history for each and generate an artificial correct future. To do so, a lateral deviation and/or an orientation deviation are added to the trajectory at some time in the past history, with a varying amplitude, deviation and recovery time.
The deviation is always introduced in the past, since the network should not be trained to induce deviations but to recover from them. To get the corresponding input image, cameras are added and recorded to emulate positions of the ego vehicle with lateral and angular offset. We choose to add 4 lateral positions every 0.2 meters on each side of the car with 3 possible orientations for the cameras: facing front, facing a point 10 meters in front of the car and the symmetric. Figure~\ref{fig:cor_beh} illustrates the simulated deviations and the corresponding recoveries, from an existing trajectory. Results from Section~\ref{sec:results} show that even if the model is simple, this label augmentation brings considerable robustness to the system at online test time.

\begin{table*}[ht]
 \centering
 \small
 \begin{tabulary}{\textwidth}{L*{3}{C}|*{3}{C}|*{3}{C}|*{3}{C}}
   \toprule
   & \multicolumn{3}{c}{Training} & \multicolumn{3}{c}{New weather} & \multicolumn{3}{c}{New town} & \multicolumn{3}{c}{New town + weather} \\
  Task & CIL & CAL & Ours & CIL & CAL & Ours & CIL & CAL & Ours & CIL & CAL & Ours\\
  \midrule
  Straight & 95 & \textbf{100} & \textbf{100} & 98 & \textbf{100} & \textbf{100} & \textbf{97} & 93 & \textbf{97} & 80 & 94 & \textbf{98} \\
  One turn & 89 & 97 & \textbf{100} & 90 & 96 & \textbf{100} & 59 & 82 & \textbf{87} & 48 & 72 & \textbf{90} \\
  Navigation & 86 & 92 & \textbf{99} & 84 & 90 & \textbf{100} & 40 & \textbf{70} & 58 & 44 & 68 & \textbf{74} \\
  Nav. dynamic & 83 & 83 & \textbf{98} & 82 & 82 & \textbf{100} & 38 & 64 & \textbf{68} & 42 & 64 & \textbf{82} \\
  \bottomrule
 \end{tabulary}
 \caption{Success rate comparison (in \% for each task and scenario, more is better) with baselines \cite{codevilla2018end, sauer2018conditional}}
 \label{tab:bench_success}
\end{table*}

\begin{table*}[ht]
\small
 \begin{center}
 \begin{tabulary}{\textwidth}{L*{3}{C}|*{3}{C}|*{3}{C}|*{3}{C}}
   \toprule
   & \multicolumn{3}{c}{Training} & \multicolumn{3}{c}{New weather} & \multicolumn{3}{c}{New town} & \multicolumn{3}{c}{New town + weather} \\
   Infraction type & CIL & CAL & Ours & CIL & CAL & Ours & CIL & CAL & Ours & CIL & CAL & Ours\\
   \midrule
   Opposite lane & 33.4 & 6.7 & \textbf{75.45} & 57.3 & \textbf{$>$ 60} & \textbf{$>$35.7} & 1.12 & \textbf{2.21} & 1.45 & 0.78 & \textbf{2.34} & 1.24  \\
   Sidewalk & 12.9 & 6.1 & \textbf{37.73} & \textbf{$>$57} & 6.0 & \textbf{$>$35.7} & 0.76 & 0.88 & \textbf{1.53} & 0.81 & 1.34 & \textbf{2.01}  \\
   Collision: static & \textbf{5.38} & 2.5 & 3.97 & 4.05 & 6.0 & \textbf{11.9} & 0.40 & 0.36 & \textbf{0.46} & 0.28 & 0.31 & \textbf{0.44}  \\
   Collision: car & 3.26 & 12.1 & \textbf{$>$75} & 1.86 & \textbf{$>$60} & \textbf{$>$35.7} & 0.59 & 2.04 & \textbf{5.12} & 0.44 & 1.38 & \textbf{16.15}  \\
   Collision: ped. & 6.35 & \textbf{30.3} & 15.09 & 11.2 & \textbf{$>$60} & 8.93 & 1.88 & \textbf{26.49} & 3.25 & 1.41 & \textbf{6.72} & 2.31  \\
   \bottomrule
 \end{tabulary}
 \captionsetup{justification=centering}
 \caption{Infraction distance comparison (in km between infractions, more is better), with baselines \cite{codevilla2018end, sauer2018conditional}. Distances with $>$ indicate that no infraction was encountered during the whole benchmark.}
 \label{tab:bench_dist}
 \end{center}
\end{table*}

In the end, all recordings joined together sums up to 15 hours of ego vehicle driving, strictly respecting the benchmark training conditions (predetermined weathers and map). About 20\% of the episodes were recorded with the ego car equipped with the 25 cameras, and all the tracks of these episodes are artificially deviated in the final dataset. The dataset on Town~01 is split into two parts only: training (90\%) and validation (10\%), testing is done online in the simulator on both maps of CARLA. We chose not to record any data on Town~02 map, even for validation of the models so it remains a pure test for the trajectory prediction.


%
%
%

\section{Results}
\label{sec:results}

\subsection{Metrics}

The CARLA benchmark (first introduced in \cite{codevilla2018end}) is used to quantify the network performance in terms of ego vehicle trajectory prediction, and compare it to the state of the art. The benchmark regroups navigation tasks of increasing difficulty: going in a straight line, taking one turn, and navigation task with several turns without or with other road users. It can be performed on the training town (Town~01) or on the test town (Town~02). The benchmarks metrics record the percentage of successfully performed tasks, which means that the vehicle reached its destination, and the distance between two infractions (driving on the wrong side of the lane, hitting something etc). In accordance with the state of the art reported results, we compare the results for Town~01 and train weathers (\textit{training}), Town~01 and test weathers (\textit{new weather}), Town~02 and train weathers (\textit{new town}) and Town~02 and test weathers (\textit{new town + weather}). Even if the benchmark takes place in the town where the training data is recorded, this is not the training data: not only this is an online test, but also the input data is not the same.

Note that the default timeout before considering an experiment failed is multiplied by 3. This is necessary because the default is calculated based on the total path distance and since our network must stop at red lights, this drastically reduces the average mean speed of the controlled car, especially in Town~02 where traffic lights can be very close to each other. This adaptation should not impact the benchmark results, the average speed while not stopped is close to the urban driving speed of Carla's autopilot (20kph).

To provide quantitative results on the neighbors trajectories prediction, we provide metrics on the trajectories from our test set. Mean Average Error (MAE) between predicted and ground truth trajectory points is used to quantify the prediction error. The MAE is computed on the whole future trajectory, but also measured for the specific points at t=2s. The goal is to differentiate between global error, short and long term prediction. The same metrics are also provided offline for the ego vehicle prediction.

\subsection{Benchmark quantitative results}

We compare our results to two different baselines: Conditional Imitation Learning (CIL) \cite{codevilla2018end} and Conditional Affordance Learning (CAL) \cite{sauer2018conditional}. Both have a structure similar to our implementation, except that CIL produces instantaneous commands and CAL produces affordances which are then given to a controller. To control the car in the simulation, a simple PID controller was used to transform trajectory predicted by the network into a car command. The speed is capped to 20kph maximum, and the control could probably be improved (filtering or smoothing), which could lead to a slight improvement on the results.

Table~\ref{tab:bench_success} summarizes the comparison for success rates on the benchmark tasks. Our method outperforms CIL on practically all tasks and all conditions, and the difference is more significant for tasks harder than the straight line. We are competitive with CAL: we equal or outperform on training conditions, and outperform CAL on the most general test conditions. We note that for the hardest tasks, dynamic navigation in new town, our method performs better when other road users are present in the simulation. In fact this situation is closer to the training data and the network must have learned to partially rely on the other vehicles position to predict trajectories.
Our method also outperforms CIL, except for the static collision infraction. The gap is especially visible in tasks related to steering (opposite lane and sidewalk). This is probably due to the label augmentation which is designed to keep the vehicle in the nominal trajectory. For test cases on car collision, our algorithm also performs very well, which is certainly due to the neighbors positions input. Note that our approach is significantly weaker for pedestrian infractions, even if it still is comparable to CIL. This is probably linked to the fact that there were few pedestrians in our training base.

\begin{table}
 \centering
 \begin{tabulary}{\columnwidth}{p{1.5cm}*{3}{p{0.9cm}}C}
 \toprule
  & Training & New weather & New town & New town + weather \\
  \midrule
  Ratio (\%) & 3.0 & 1.8 & 21.9 & 25.3 \\
  Total & 436 & 436 & 283 & 170 \\
  \bottomrule
 \end{tabulary}
 \caption{Evaluation of the number of red light runs of the network: percentage of red lights run over all encountered traffic lights count (dynamic navigation task of the benchmark, lower is better)}
 \label{tab:red_lights}
\end{table}

Table~\ref{tab:red_lights} shows a quantitative evaluation of the traffic lights infractions. We report the ratio of red lights run over all encountered traffic lights for the dynamic navigation task. Note that this number is affected by the traffic light states, because the network cannot run a light that is green, and the states encountered vary with the vehicle speed, starting positions, traffic state etc.

\subsection{Ablation studies}

The impact of our data augmentation is measured in Table~\ref{tab:bench_success_da}. Three scenarios are compared, in Town~01 : no label augmentation (only randomization is introduced), partial label augmentation (performed on only one of the 4 train weathers), and full (performed on all train weathers). Table~\ref{tab:mae_comparaison} compares the MAE obtained on  Town~01 train and validation data weathers.

\begin{table}[h]
\small
\centering
 \begin{tabulary}{\columnwidth}{p{2cm}*{3}{C}|*{3}{C}}
  \toprule
   & \multicolumn{3}{c}{Train} & \multicolumn{3}{c}{Validation} \\
  Augmentation & None & Part & Full & None & Part & Full \\
  \midrule
  Straight & 70 & 100 & 100 & 54 & 100 & 100 \\
  One turn & 21 & 100 & 100 & 14 & 100 & 100 \\
  Navigation & 16 & 93 & 99 & 12 & 96 & 100 \\
  Nav. dynamic & 14 & 97 & 98 & 4 & 100 & 100 \\
  \bottomrule
 \end{tabulary}
\caption{Comparative success rates (in \%) without, with partial and full data augmentation in Town~01  (None, Partial, Full) for train and validation weathers}
\label{tab:bench_success_da} 
\end{table}

The comparison in Table~\ref{tab:bench_success_da} confirms that label augmentation is critical to the performance of the vehicle control: there is a significant gap when introducing the augmentation, then a slight improvement when we deploy it on all conditions. It is interesting to note that the effect is much more noticeable on complex navigation tasks, where the maneuvers are more complex so that errors accumulate quicker. However, the offline quantitative evaluations of the MAE from Table~\ref{tab:mae_comparaison} are almost identical. This proves that online test is the real significant indicator for IL when it is used for active control. Note that this is in line with the findings of \cite{codevilla2018offline}, which highlights that the correlation between offline metrics and online performance is weak.

\begin{table}[h]
\small
\centering
 \begin{tabulary}{\columnwidth}{p{2.0cm}*{3}{J}|*{3}{J}}
  \toprule
  Weathers & \multicolumn{3}{c}{Train} & \multicolumn{3}{c}{Validation} \\
  Augmentation & None & Part & Full & None & Part & Full \\
  \midrule
  Ego & \textbf{0.05} & 0.06 & 0.06 & \textbf{0.09} & \textbf{0.09} & 0.10 \\
  Ego 2s & \textbf{0.11} & 0.12 & 0.13 & 0.23 & 0.23 & 0.23 \\
  Neighbors & 0.10 & 0.10 & 0.10 & 0.23 & 0.22 & \textbf{0.21} \\
  Neighbors 2s & 0.22 & 0.22 & \textbf{0.21} & 0.54 & 0.54 & \textbf{0.51} \\
  \bottomrule
 \end{tabulary}
\caption{Comparison of MAE on train and validation data (in m), with none, partial and full data augmentation (None, Partial, Full), less is better}
\label{tab:mae_comparaison} 
\end{table}

Table~\ref{tab:mae_comparaison} also shows that the error is greater for the neighbors than for the ego. This is not surprising, because the ego trajectory has the additional information of the navigation command, whereas neighbor trajectories can be ambiguous at intersections for a limited time, in which case there is an error on the future trajectory. In the case of neighbours prediction, the measure is more relevant because there is no online control loop for the neighbors.

\begin{table}[h]
 \small
 \centering
 \begin{tabulary}{\columnwidth}{J*{2}{C}|*{2}C}
 \toprule
   & \multicolumn{2}{c}{Train} & \multicolumn{2}{c}{Validation} \\
  With neighbors & Yes & No & Yes & No \\
  \midrule
  Navigation (\%) & 58 & \textbf{84} & 74 & \textbf{92}\\
  Nav. dynamic & 68 & \textbf{81} & 82 & \textbf{92} \\
  Opposite lane (km) & \textbf{1.45} & 0.62 & \textbf{1.24} & 0.19\\
  Sidewalk & \textbf{1.53} & 0.32 & \textbf{2.01} & 0.16 \\
  Static & 0.46 & \textbf{1.26} & 0.44 & \textbf{0.60} \\
  Car & \textbf{5.12} & 0.25 & \textbf{16.15} & 0.19 \\
  Pedestrian & \textbf{3.25} & 0.25 & \textbf{2.31} & 0.25 \\
  \bottomrule
 \end{tabulary}
 \caption{Ablation study of the neighbor prediction for Town~02 train and validation weathers (for navigation tasks)}
 \label{tab:ablation_neighb}
\end{table}

Table~\ref{tab:ablation_neighb} illustrates the influence of the neighbors trajectory prediction on the global performance. In this ablation, the loss on neighbors is removed, but the proximity map is kept, so there is no backpropagation on the neighbors encoding and decodings. Very interestingly, predicting the neighbors actually reduces the percentage of tasks completed, but performs significantly worse for all infractions except static collisions. It appears that adding the neighbors prediction makes the ego prediction more compliant to traffic rules. This opens new research directions to investigate how mixing neighbors and ego trajectory prediction impacts performances.

\begin{table}[h]
\centering
\small
 \begin{tabulary}{\columnwidth}{*{2}{C}|*{2}{C}|*{2}{C}|*{2}{C}|*{2}{C}}
  \toprule
  \multicolumn{2}{c}{None} & \multicolumn{2}{c}{Medium} &  \multicolumn{2}{c}{High} & \multicolumn{2}{c}{Neighbors} & \multicolumn{2}{c}{YOLO} \\
  train & test & train & test & train & test & train & test & train & test \\
  \midrule
  68 & 82 & 69 & 80 & 63 & 76 & 74 & 74 & 67 & 78 \\
  \midrule
  \multicolumn{10}{c}{3D bounding box noise standard deviation (m)} \\
  x & y & x & y & x & y & x & y & \multicolumn{2}{c}{(Axis)} \\
  0.0 & 0.0 & 0.1 & 0.05 & 0.3 & 0.15 & 0.1 & 0.05 & \multicolumn{2}{c}{(Ego)} \\
  0.0 & 0.0 & 0.1 & 0.05 & 0.3 & 0.15 & 1.0 & 0.5 & \multicolumn{2}{c}{(Neighb.)} \\
  \bottomrule
 \end{tabulary}
\caption{\% success rate for navigation with varying noise (train \& test weathers) and YOLO images bounding boxes}
\label{tab:bench_noise} 
\end{table}

A final ablation study has been done on the influence of noise on vehicle positions. In this all other evaluations, the ground truth positions are used, but this is not available in the real world and could constitute a bias in the comparison with state of the art. To prove the robustness, a Gaussian white noise of increasing deviation is added to the positions of the objects, both at training and testing time. In parallel, the image bounding boxes, which originally are a projection of the objects positions in the image, are replaced by the output of a YOLO \cite{redmon2016yolo9000} detector trained on CARLA images. For time constraints, the study is conducted only for the success rate of the dynamic navigation task. Table~\ref{tab:bench_noise} illustrates this: even for high levels of noise and using an image detector, the performance drop is limited. This proves that any reasonably robust object detector could be used instead of the ground truth position.

\subsection{Qualitative evaluation}

\begin{figure}[t]
 \centering
 \includegraphics[width=0.95\columnwidth]{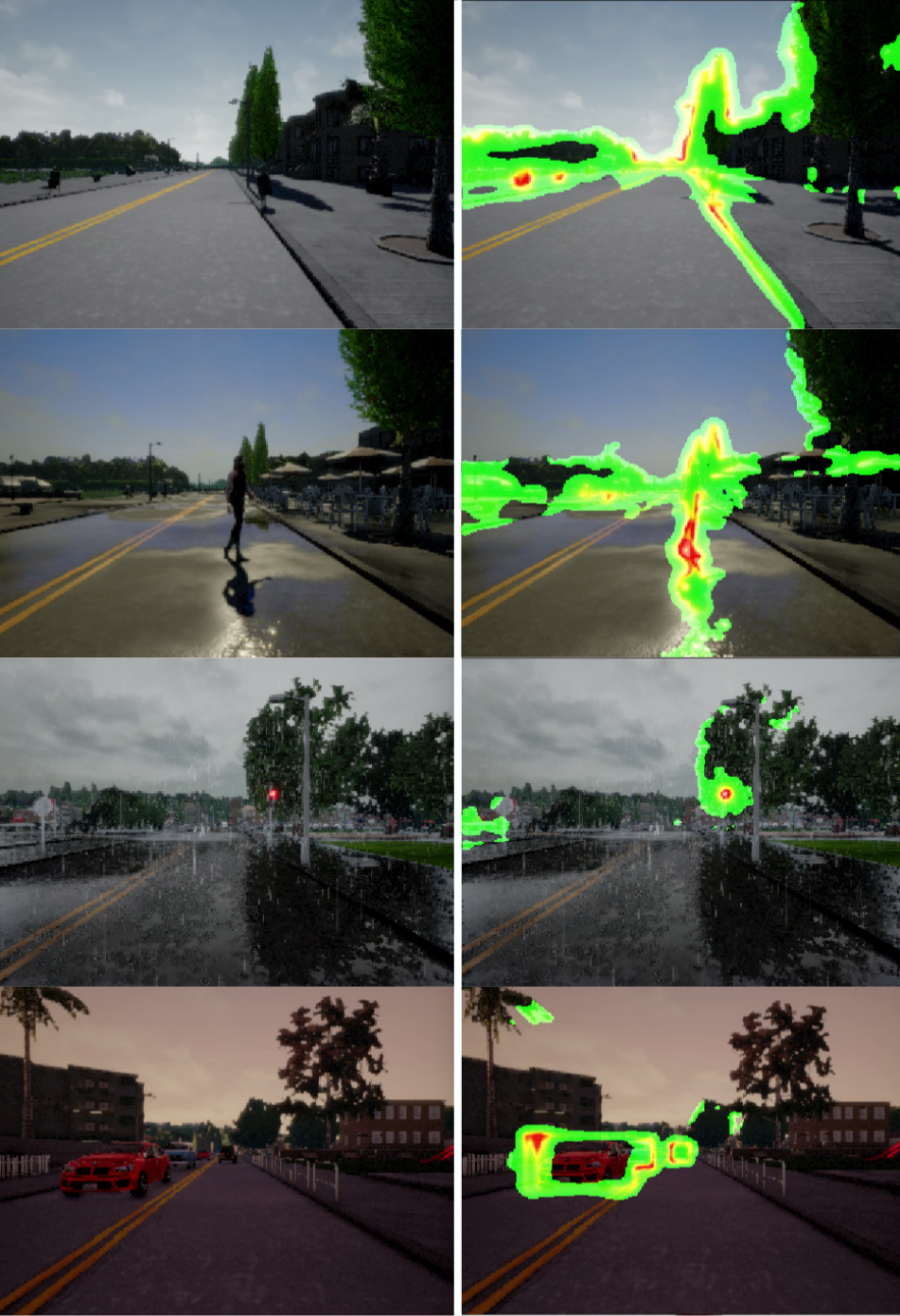}
 \caption{VisualBackprop (right) for ego trajectory prediction and corresponding image (left), best viewed in color. The heatmap overlay represents area of activation(green is low and red is high). From top to bottom: ego car too close to the sidewalk, pedestrian crossing, before stopping at red light, dense traffic}
 \label{fig:backprop}
\end{figure}

To qualitatively explain the outputs of the network, we rely on VisualBackProp \cite{bojarski2018visualbackprop}, which highlights the image pixels which contributed the most to the final results. Figure~\ref{fig:backprop} shows typical examples of the backpropagation. Vehicles are predominant, which is not suprising because their boxes are provided as input. More interestingly, the VisualBackprop also highlights lane markings and curbs which are relevant for lateral positioning, but mostly when there is a significant lateral deviation. Traffic lights with the current color light and pedestrians are also highlighted when relevant. This is consistent with the fact that this information can only be found in the image, and not in the position history, and shows that it is possible to learn a link between the image and the metric space.

\begin{figure}
 \centering
 \includegraphics[width=0.9\columnwidth]{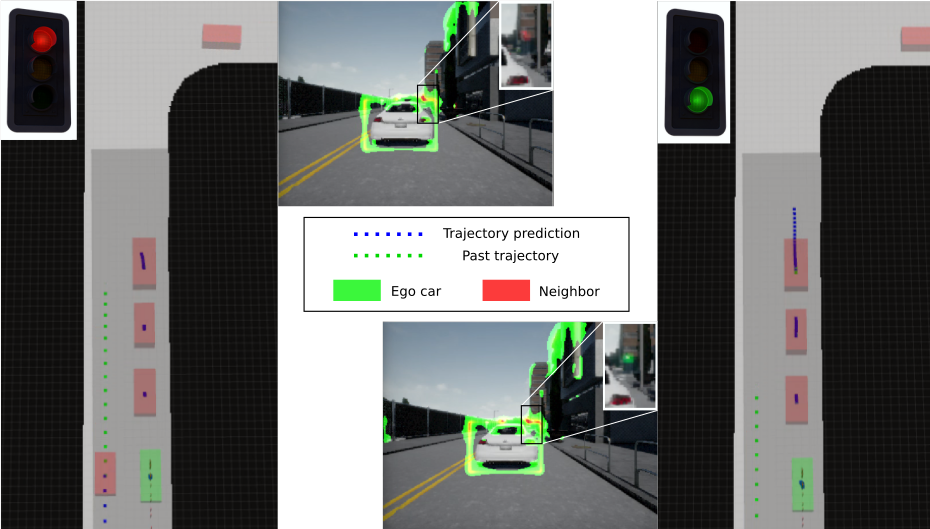}
 \caption{Image and predicted trajectories at red light (top), right after light turns green (bottom). The vehicle at the front of the queue has a restart trajectory predicted, but the ego vehicle is still predicted to be stopped for the next 2s.}
 \label{fig:traffic_light}
\end{figure}

The autopilot in CARLA respects traffic rules, in particular traffic lights. They are taken into account by the trained network, not only for the ego vehicle, but also for the neighbors. A typical situation is a queue at a red light, with at least one vehicle standing before the ego vehicle. While the light is red, all vehicles are predicted stopped. When the light turns green, the trajectory of the first vehicle in the line predicts a restart while the others are still stopped. Then the vehicles are predicted to restart progressively, one after the other. Figure~\ref{fig:traffic_light} is a typical example of how position and image information are correlated, and how the networks takes vehicle interactions into account.

The supplementary material contains a video of the network driving around CARLA Town~01, with the VisualBackprop and the vehicles trajectory predictions (on top view and projected on the ego vehicle frontal camera).

\section{Conclusion}

In this work, we have proposed a hybrid end-to-mid neural network to predict vehicle trajectories in CARLA urban simulated environment. The network integrates positional and image information, learns vehicle interactions and extracts relevant data from image. It is applied to a navigation scenario, and produces significant improvement over previous state of the art on the CARLA benchmark. This shows that a mix of high and low level data, together with auxiliary tasks, bring performance to imitation learning. This work also highlights the impact of label augmentation: adding artificial data helps reduce the gap between train and test distribution and increases the performance drastically.

In future work, this framework could be applied on real data.
To make the network truly end-to-mid, the object detection input could be integrated inside the network, with an auxiliary goal to be trained on object detection or a similar task. Investigations on the influence of additional predictions would bring insights for the structuring of end-to-end networks.

{
\small

}

\end{document}